\documentclass[11pt]{article}

\usepackage[preprint]{acl}

\usepackage{times}
\usepackage{latexsym}
\usepackage{booktabs}
\usepackage{amssymb}
\usepackage{array}
\usepackage[table]{xcolor}
\usepackage{subcaption}
\usepackage{fontawesome5}

\usepackage[T1]{fontenc}

\usepackage[utf8]{inputenc}

\usepackage{microtype}

\usepackage{inconsolata}

\usepackage{graphicx}

\newcommand{\drafttrace}{\textsc{DraftTrace}}

\title{\drafttrace{}: A Multi-View Analytics Environment for AI-Integrated Writing}

\author{
\textbf{Divyansh Chandarana}\footnotemark[1]
\quad
\textbf{Sandipan De}\footnotemark[1]
\quad
\textbf{Vivek Gupta}
\\[2pt]
Arizona State University
\\[4pt]
\href{https://draftrace.com/}{\faPlayCircle\ Demo}
\quad
\href{https://www.youtube.com/watch?v=gCpNzqm75FU}{\faVideo\ Video}
\\[3pt]
\texttt{\{dchanda1, sandipan, vgupt140\}@asu.edu}
}

\begin{document}

\maketitle

\begingroup
\renewcommand{\thefootnote}{\fnsymbol{footnote}}
\footnotetext[1]{Contributed equally.}
\endgroup

\begin{abstract}
Generative AI has changed how students produce writing assignments. The final artifact is no longer sufficient to understand the process through which it was produced. We introduce \drafttrace{}, a writing environment that jointly captures three complementary views of writing: the final product, the writing process and interactions with an integrated AI-assistant. \drafttrace{} reconstructs how a document develops over time and organizes these signals into submission-, longitudinal-, and class-level analytics for instructors. We deployed \drafttrace{} in a graduate NLP course with 81 students and compared their sessions with LLM-generated responses entered by automated tools and with copy-typed responses. While product measures distinguish differences in text formulation, process measures distinguish differences in how text is entered. Considering both views together helps characterize cases such as copy-typing. Interaction traces show that students use the assistant differently across stages of writing: to clarify the question at an early stage and to verify answers at a later stage. A preliminary instructor survey highlights the importance of multi-view writing analytics and their interpretability.
\end{abstract}

\section{Introduction}
Generative AI (GenAI) has changed not only the artifacts students submit as part of written assignments, but also the processes through which those artifacts are produced. Students can use AI assistance at different stages of an assignment: to understand a concept, brainstorm ideas, construct an outline, revise existing writing, co-write portions of a response, or generate a complete response. Traditionally, although students draw on textbooks, scholarly literature, online resources, or other external materials, producing a coherent response generally required them to engage with the material. GenAI has substantially changed this relationship between the final product and students' engagement, and has made it increasingly difficult for instructors to determine whether and how the writing activity contributed to student learning.

Prior work has studied the writing process by capturing measures like writing speed, pauses, bursts, revisions and deletion events during text production. These measures have been used to investigate their relationships with underlying cognitive processes. However, such behavioral signals do not map uniquely to specific cognitive activities; for example a pause can be associated with planning, reflection, or revision. GenAI introduces an additional source of ambiguity into these observations. A pause may now also correspond to interaction with AI system.

These changes call for a reconsideration of how writing activities are characterized, measured and interpreted in AI-mediated settings. We argue that understanding of writing requires considering three complementary views: the \emph{product view}, which characterizes the final artifact; the \emph{process view}, which captures how the artifact develops over time; and the \emph{interaction view}, which captures how the student engages with the AI during that process. Existing tools provide different subset of these capabilities, including final product analysis, revision histories, and integrated AI assistance. However, these capabilities are treated as separate signals rather than a complementary views of the same writing activity.

To support these integrated perspective, we introduce \drafttrace{} a writing environment that jointly capture the writing process, characterize the resulting product, and records students' interactions with an integrated AI assistant. \drafttrace{} organizes these complementary sources of information into instructor-facing views at multiple levels of granularity.

\section{Background and Related Work}
\subsection{Writing Analytics}
Writing analytics has used characteristics of completed text for applications ranging from automated essay scoring to automated feedback generation. \citet{ke2019automated} survey
product-level measures spanning style, relevance, organization, cohesion, and coherence, among other dimensions of writing quality. More recently, \citet{pande_providing_2026} develop a pipeline that utilizes stylometric analytics alongside a large language model (LLM) to generate feedback for student writing at scale.

Writing research has also used keystroke events to study the writing process. \citet{guo_modling_2018} show that longer writing time and shorter and less variable within-word keystroke intervals are associated with higher essay scores. This finding is consistent with the view that fluency in lower-level transcription processes may free cognitive resources for higher-level composition. \citet{schaller-etal-2026-keyscore} further demonstrate that keystroke-derived features can provide predictive signals of essay quality during the early stages of composition, before sufficient textual content is available for conventional product-based scoring. At the same time, \citet{babalola_stability_2026} find that many writing-process features exhibit variability across tasks and contexts, including grade level, academic proficiency, and school setting.

\subsection{AI-Mediated Writing}
Studies of AI-mediated writing have examined how differences in AI access and interaction behavior relate to students' writing processes. While \citet{christenson-etal-2026-effects} measure how ownership perception of the student changes based on the frequency of LLM access, \citet{minju_characterizing_2026} categorize the types of LLM use and examine their association with student performance. Complementing these studies \citet{yang_modifying_2026} examine how students process AI-generated assistance, distinguishing between suggestions that are rejected, accepted with modifications, or accepted unchanged. Collectively, these findings highlight substantial variation in how student engage with and use AI during writing.

Recent work has started to combine these perspectives. \citet{he_beyond_2025} combine process and product information to improve writing assessment. In AI assisted writing, \citet{chen_Beyond_chat} incorporate interaction traces to characterize how students integrate AI into their writing. While these studies combine subsets of signals for specific analytical tasks, collectively they suggest the value of considering product, process, and interaction together to provide a comprehensive view of student writing.

\section{\drafttrace{}}
The design of \drafttrace{} is guided by four goals that determine what information is captured, how assistance is provided, how resulting information is presented and how the platform fits within existing educational workflows.

\paragraph{Integrated view of writing.} Process, product, and interaction information is captured and aligned as complementary views of the writing.

\paragraph{Configurable AI support.} Rather than treating AI access as allowed or forbidden, instructors can configure the level of AI assistance according to the assignment and learning objectives.

\paragraph{Longitudinal and class-level analytics.} Writing analytics are organized to compare a student's submissions over time and with broader class-level patterns.

\paragraph{Low-friction access and integration.} The writing environment operates independently or integrates with existing Learning Management Systems to allow students and instructors to use the platform within their normal workflows without requiring separate procedures.

Existing writing and educational platforms provide different subsets of capabilities to these design goals. Table~\ref{tab:system-comparison} compares \drafttrace{} with representative systems across these capabilities.









\begin{table}[!ht] 
\centering 
\small 
\setlength{\tabcolsep}{2.2pt} 
\renewcommand{\arraystretch}{1.4} 
\begin{tabular}{
>{\raggedright\arraybackslash}m{1.5cm}
@{\hspace{12pt}}
cccccccc
} 
\toprule 
\textbf{System} 
& \rotatebox{75}{\textbf{Proc.}} 
& \rotatebox{75}{\textbf{Prod.}} 
& \rotatebox{75}{\textbf{Interact.}} 
& \rotatebox{75}{\textbf{Assist.}} 
& \rotatebox{75}{\textbf{Config.}} 
& \rotatebox{75}{\textbf{Playback}} 
& \rotatebox{75}{\textbf{Analytics}} 
& \rotatebox{75}{\textbf{Editor}} \\ 
\midrule 
 
DraftTrace 
    & \checkmark & \checkmark & \checkmark 
    & \checkmark & \checkmark 
    & \checkmark & \checkmark & \checkmark \\ 
 
Turnitin Clarity 
    & \checkmark & \checkmark & -- 
    & \checkmark & -- 
    & \checkmark & \checkmark & \checkmark \\ 
 
Documark 
    & \checkmark & \checkmark & -- 
    & -- & -- 
    & \checkmark & \checkmark & \checkmark \\ 
 
Grammarly Authorship 
    & \checkmark & \checkmark & -- 
    & -- & -- 
    & \checkmark & \checkmark & \checkmark \\ 
 
Draftback 
    & \checkmark & -- & -- 
    & -- & -- 
    & \checkmark & \checkmark & \checkmark \\ 
 
GPTZero 
    & -- & \checkmark & -- 
    & -- & -- 
    & -- & -- & -- \\ 
 
Pangram 
    & -- & \checkmark & -- 
    & -- & -- 
    & -- & -- & -- \\ 
 
\bottomrule 
\end{tabular} 
 
\caption{Comparison of capabilities across representative writing 
platforms. Proc.: process view; Prod.: product view; Interact.: 
interaction view; Assist.: integrated AI assistant; Config.: configurable 
AI support; Editor: integrated writing environment.} 
\label{tab:system-comparison} 
\end{table}

\subsection{System Architecture}
\begin{figure*}[ht]
    \setlength{\fboxrule}{0.1pt}
    \fbox{\includegraphics[width=0.95\linewidth]{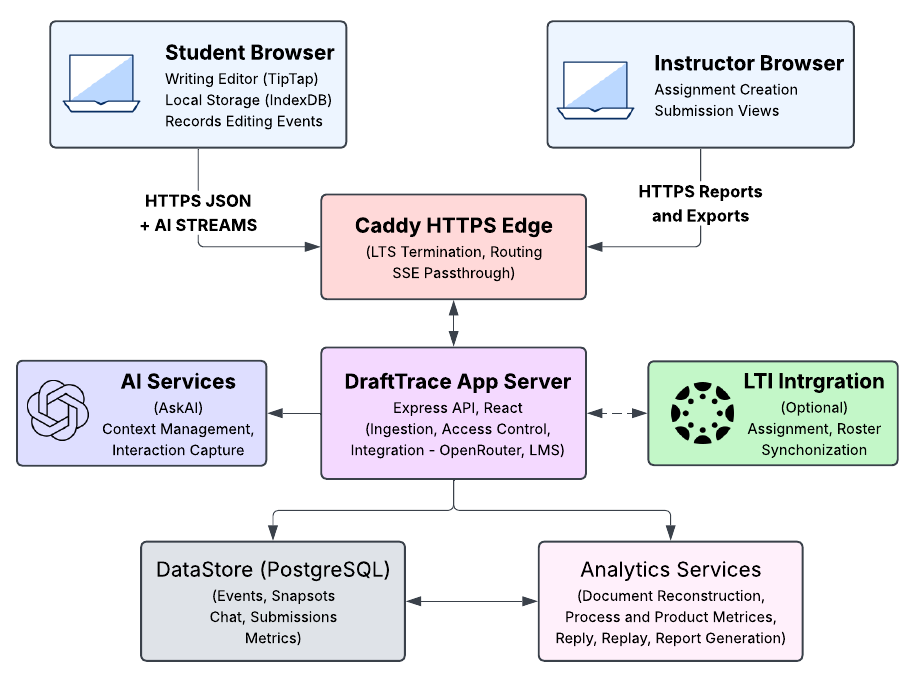}}
    \caption{Overview of the \drafttrace{} system architecture for capturing, reconstructing, and analyzing student writing and AI interactions.}
    \label{fig:architecture}
\end{figure*}

\drafttrace{} is a web application with a React front end, Express back end, and a PostgreSQL store (Figure~\ref{fig:architecture}). The editor is built on TipTap/ProseMirror. The browser and server share the same document schema and analysis modules, so the writing playback shown to instructors and the metrics computed on the server are derived from identical logic.

\paragraph{Process capture.}
\drafttrace{} records each change to the document as an event with a timestamp, associated text, and its input channel: typing, paste, drag-and-drop, undo/redo, or insertion from the integrated assistant. Switching away from the editor is also logged. Complete document snapshots are stored periodically to bound replay cost.

To avoid data loss on unreliable classroom networks, events are queued in the browser's IndexedDB and then uploaded small batches. Each event carries a unique sequence number, making retries idempotent.  The events are discarded locally only after the server acknowledges them. For timed assignments, deadlines are enforced on the server, and incomplete sessions are submitted automatically when time expires.

\paragraph{Product reconstruction and provenance.}
After submission, \drafttrace{} replays the recorded events to rebuild the document and labels each character by how it entered: typed, inserted from the built-in assistant, pasted with or without a cited source, or unknown. These labels are summarized per paragraph and for the whole submission, alongside process measures such as bursts, pauses, and revisions.

\paragraph{Interaction capture.}
The integrated assistant is proxied through the server rather than called from the browser. Each student prompt is persisted. Responses are streamed to the student and saved to the session transcript which is visible to the instructor. Text inserted from
the assistant is tagged at the transaction level, linking the interaction view to the process and product views. Instructors can enable or disable the assistant per assignment, and a system prompt instructs it to support understanding rather than write answers.

\subsection{Instructor Interface}
\begin{figure}[ht]
    \setlength{\fboxrule}{0.1pt}
    \fbox{\includegraphics[width=0.95\columnwidth]{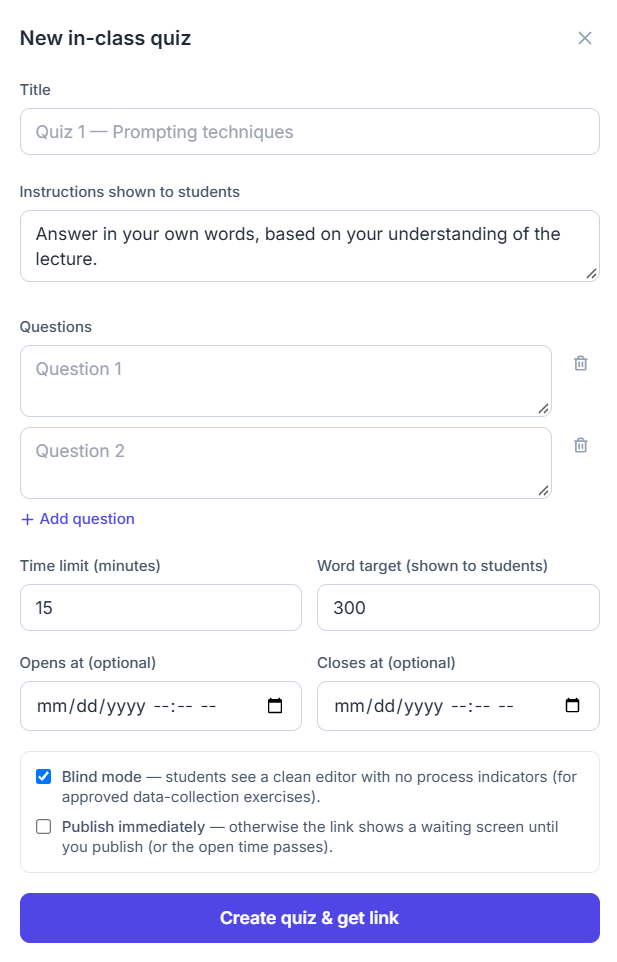}}
    \caption{Instructor interface for creating an assignment.}
    \label{fig:quiz-creation}
\end{figure}

\begin{figure}[ht]
    \setlength{\fboxrule}{0.1pt}
    \fbox{\includegraphics[width=0.95\columnwidth]{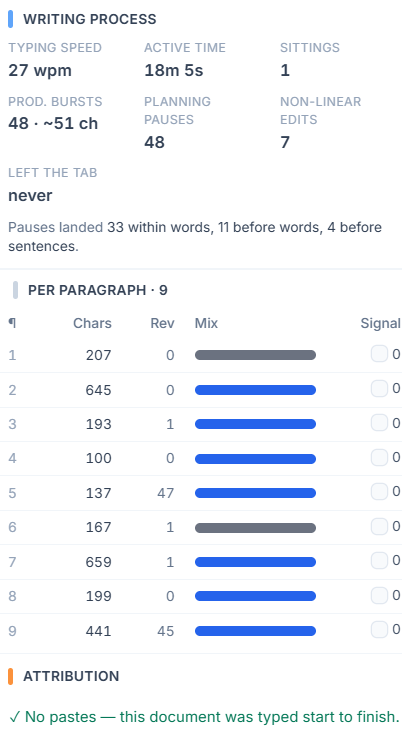}}
    \caption{Writing-process view summarizing measures captured during an individual writing session.}
    \label{fig:rpt-process-view}
\end{figure}

The instructor interface supports two primary workflows - to create quizzes and to review the resulting analytics. Instructors can configure the questions, instructions, time and length constraints (Figure~\ref{fig:quiz-creation}). For each submission, \drafttrace{} summarizes writing-process measures (Figure~\ref{fig:rpt-process-view}) and provides a chronological timeline of the writing session (Figure~\ref{fig:rpt-activity-timeline}). Additional class-level and longitudinal views are provided in Appendix~\ref{sec:additional-interfaces}.

\begin{figure}[ht]
    \setlength{\fboxrule}{0.1pt}
    \fbox{\includegraphics[width=0.95\columnwidth]{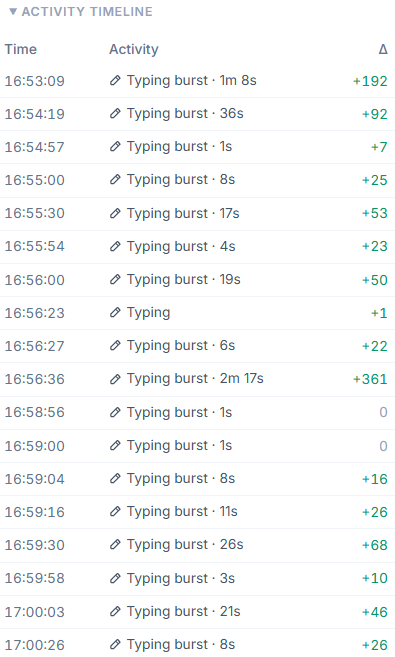}}
    \caption{Activity timeline showing the chronological development of an individual writing session.}
    \label{fig:rpt-activity-timeline}
\end{figure}

\subsection{Student Interface}
\begin{figure}[ht]
    \setlength{\fboxrule}{0.1pt}
    \fbox{\includegraphics[width=0.95\columnwidth]{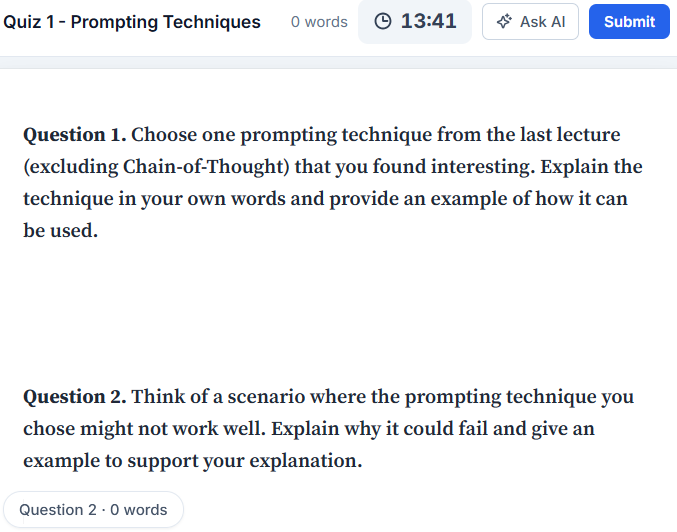}}
    \caption{Writing
    environment for completing the assigned task.}
    \label{fig:student-write}
\end{figure}

\begin{figure}[!ht]
    \setlength{\fboxrule}{0.1pt}
    \fbox{\includegraphics[width=0.95\columnwidth]{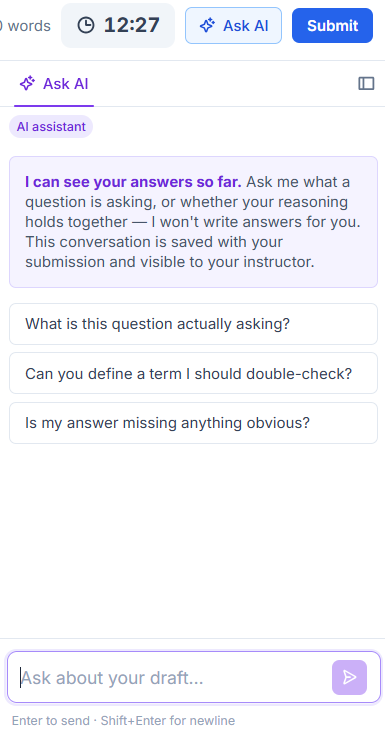}}
    \caption{Integrated
    AI assistant.}
    \label{fig:student-ask-ai}
\end{figure}

The student interface provides an environment for completing the assignment task. Student compose their responses directly in the editor (Figure~\ref{fig:student-write}). When enabled by instructor, students can access the integrated AI assistant from within the same environment (Figure~\ref{fig:student-ask-ai}).

\section{Case Study}
\subsection{Data Collection}
We collected writing traces under three settings that vary how the written response is formulated and entered into the writing environment: classroom writing, automated writing, and copy-typing.

\paragraph{Classroom Writing.} We used \drafttrace{} during a writing activity in graduate-level Natural Language Processing course and collected responses from 81 students who were present during the session. At the beginning of the activity, students were briefly introduced to the writing environment and its available features. Students were given 15 minutes to answer two related questions based on the materials covered in a previous lecture. They were allowed to consult lecture slides and use the AI assistant integrated within \drafttrace{}, but were instructed not to use external AI tools. 

The first question asked the students to select a prompting technique and explain it in their own words and provide examples of its use. The second question asked the students to identify a scenario in which the selected technique might not work well and explain why. While all students followed the same general task structure, they could select different prompting techniques and construct their own examples, allowing variation in the content of the responses.

\paragraph{Automated Writing.}
We collected an additional 81 writing sessions in which response to the same questions were generated through LLM. The generated responses were entered into \drafttrace{} using automated typing mechanism that approximates human-like text entry.

\paragraph{Copy-Typing.} To examine writing traces when text formulation and text entry are separated, we collected additional eleven sessions in which participants were allowed to use provided responses, external AI tools, or other resources while answering the same questions. Participants were instructed to manually type text into \drafttrace{}.


\subsection{Analysis}

We compare the three settings using a small set of measures from the process and product views. From the process view we use typing speed (words per minute), revisions per 100 typed characters, the number of times the student left the editor, and the fraction of elapsed time spent actively writing. From the product view we use Flesch Reading Ease and mean word length, computed on the concatenated answers to both questions. 

\begin{table}[t]
\centering
\small
\setlength{\tabcolsep}{3.5pt}
\begin{tabular}{lrrr}
\toprule
 & \textbf{Class.} & \textbf{Auto.} & \textbf{Copy} \\
 & ($n{=}81$) & ($n{=}81$) & ($n{=}11$) \\
\midrule
\multicolumn{4}{l}{\textit{Product view}} \\
Flesch Reading Ease       & 52.8 & 38.4 & 39.9 \\
Mean word length (chars)  & 4.67 & 5.18 & 5.40 \\
Mean sentence length (chars) & 130.50 & 163.17 & 147.50 \\
\midrule
\multicolumn{4}{l}{\textit{Process view}} \\
Typing speed (wpm)        & 27.5 & 49.0 & 33.0 \\
Revisions / 100 chars     & 12.5 & 2.7  & 5.4  \\
Left the editor (count)   & 14   & 0    & 0    \\
Active / elapsed time     & 0.77 & 1.00 & 1.00 \\
\bottomrule
\end{tabular}
\caption{Median values per setting. Class.: classroom writing; Auto.: automated writing; Copy: copy-typing.}
\label{tab:settings}
\end{table}

\paragraph{The product view reflects who formulated the text.}
Responses formulated by an LLM were harder to read and used longer words than classroom responses, regardless of how they were entered. Median Flesch Reading Ease was 52.8 for classroom responses, compared with 38.4 for automated and 39.9 for copy-typed responses, and median word length rose from 4.67 to 5.18 and 5.40 characters. Automated and copy-typed responses had nearly identical product measures. This is expected as the text originates from an LLM, the final artifact looks similar irrespective of how it was entered.

\paragraph{The process view reflects how the text was entered.}
Automated entry left a regular trace: a median typing speed of 49 wpm, 2.7 revisions per 100 characters, and no idle time or departures from the editor. For copy-typing we observe similar behavior: they neither paused nor left the editor as the text was already formulated. However, the participants typed more slowly (33 wpm) and revised twice as often (5.4 per 100 characters). Classroom writers, by contrast, revised most (12.5 per 100 characters), left the editor a median of 14 times, and were actively writing for 77\% of the session.

\paragraph{Neither view is sufficient alone.}
Copy-typing illustrates why the views must be read together. Its product resembles automated writing, while its process resembles human typing without the pauses and editor departures seen in the classroom. Identifying such a session requires combining both views, the integrated perspective \drafttrace{} is designed to provide.

\paragraph{The interaction view reflects when and why students sought help.}
In the classroom setting, 32 of 81 students used the integrated assistant. In total, they have issued 75 prompts (median 2 per student; 15 students asked only once). We divided each student's session into four equal quarters and assigned each prompt to a quarter based on its timestamp. Distribution of the requests across these segments is shown in Table~\ref{tab:ai-quarters}

\begin{table}[t]
\centering
\small
\begin{tabular}{lcccc}
\toprule
 & \textbf{Q1} & \textbf{Q2} & \textbf{Q3} & \textbf{Q4} \\
\midrule
Prompts                  & 11 & 17 & 21 & 26 \\
Students' first prompt   & 10 & 5  & 7  & 10 \\
\bottomrule
\end{tabular}
\caption{Assistant use by session quarter for the 32 students who used the integrated assistant.}
\label{tab:ai-quarters}
\end{table}

To examine how students used the assistant, we categorized their interactions into four types: \textit{clarification} about the question, \textit{verification} of a drafted answer, \textit{direct help} (requesting an answer or example), and \textit{collaboration} (an extended exchange that develops the response over several turns). Students whose first prompt came in the first quarter sought clarification (6 of 10) or direct help (4 of 10). Students who first used the assistant in the last quarter almost always sought verification of their answers (9 of 10). Collaboration appeared only among students who issued three or more prompts (7 of 13). Students who issued only one or two prompts generally used the assistant for either direct help or verification.

These patterns show that students use the assistant differently at different stages of the writing process. \drafttrace{} aligns the interaction view with the process timeline and provides additional context for interpreting these patterns.

\section{Instructor Survey}
To complement the classroom study, we conducted a survey\footnote{\url{https://forms.gle/jV5jeqQnwTnve7m28}} to understand instructors' current practices, their perceptions of AI-integrated writing environments, and their considerations regarding the collection and analysis of student writing data. The survey targets educators involved in teaching, assisting, or grading student work across different academic levels and disciplines.

\paragraph{Findings.}
\begin{itemize}
    \item Current practices provide limited confidence. Nine of the ten respondents reported encountering suspected inappropriate AI use at least sometimes, while only one reported being very confident in their current process for investigating such cases. This contrast suggests a gap between the frequency with which instructors encounter questions about how student work was produced and their confidence in the information currently available to investigate those questions.
    \item Writing support and visibility were valued across multiple dimensions. Writing-process visibility and final-submission analysis were each rated as very valuable by 7 of 10 respondents, while instructor-controlled AI access, AI-interaction visibility, and writing-process playback were each rated very valuable by 6 respondents.
    \item Acceptance of information collection was largely conditional. While 6 of 10 respondents considered copy/paste activity and AI-interaction history generally acceptable to collect, writing/editing activity, detailed typing patterns, and AI processing of student writing were more often viewed as acceptable only under limited circumstances.
    \item Accuracy and interpretability were identified as important considerations. Inaccurate or misleading results could discourage adoption for 9 of 10 respondents, while 4 of 10 indicated that a lack of transparency about how such systems work could affect adoption.
\end{itemize}

\section{Discussion}

While \drafttrace{} is designed as a writing analytics tool, we also view it as a research apparatus for studying how writing changes in AI-integrated environments. By collecting aligned product, process, and interaction information, the environment enables researchers to observe not only what students produce, but how their writing and use of assistance evolve during an assignment and across assignments. Beyond observation, \drafttrace{} provides an environment for adaptive experimentation. As the AI assistance is configured per assignment, researchers can vary its availability and behavior. This enables controlled studies of when assistance should be provided and what form it should take. As the system already observes writing and interaction patterns, it could be extended to adapt the level or form of assistance based on these observations, a direction we plan to explore in future work


\section{Limitations}
\drafttrace{} captures activity occurring within its writing environment and cannot observe all external resources or activities that may contribute to a response. While the captured process, product, and interaction signals provide complementary information, they do not capture every aspect of the writing process.
Additionally, our case study is limited to a single graduate-level course and a short writing activity. The writing patterns may vary across tasks, student populations, and instructional settings.

\section{Ethical Considerations}
\drafttrace{} collects fine-grained information about students' writing behavior and AI interactions beyond what is contained in a conventional final submission. Students should therefore be informed about what information is collected, how it is processed, and who has access to it. Such data, particularly writing traces and AI conversations, should be treated as sensitive educational data with appropriate access and retention controls.

\bibliography{custom}

\appendix

\section{Interfaces}
\label{sec:additional-interfaces}
\begin{figure}[t]
    \centering

    \begin{subfigure}[t]{0.92\linewidth}
        \centering
        \fbox{\includegraphics[width=0.95\linewidth]{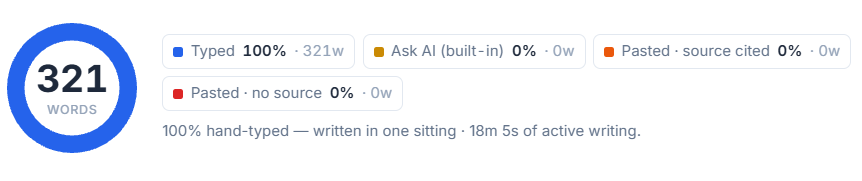}}
        \caption{Overall analytics for an individual submission.}
        \label{fig:analytics-overall}
    \end{subfigure}

    \vspace{2em}

    \begin{subfigure}[t]{0.92\linewidth}
        \centering
        \fbox{\includegraphics[width=0.95\linewidth]{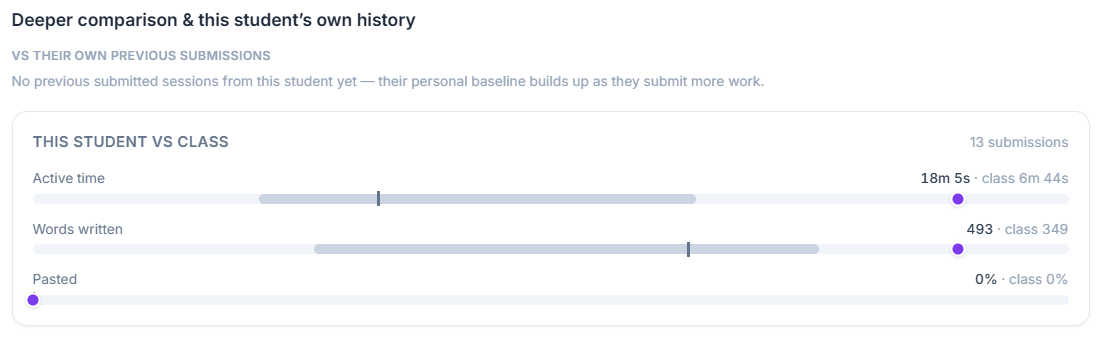}}
        \caption{Longitudinal view of a student's writing across submissions.}
        \label{fig:analytics-longitudinal}
    \end{subfigure}

    \vspace{2em}

    \begin{subfigure}[t]{0.92\linewidth}
        \centering
        \fbox{\includegraphics[width=0.95\linewidth]{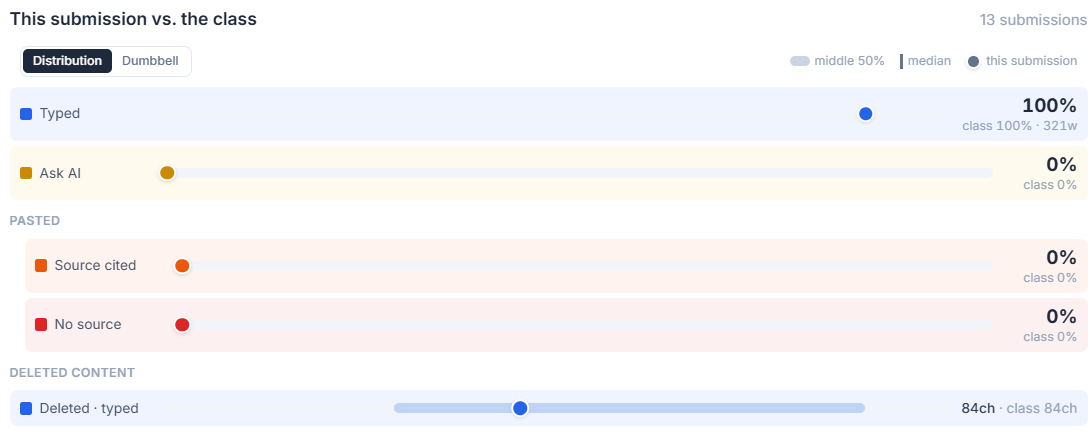}}
        \caption{Comparison of an individual submission with class-level patterns.}
        \label{fig:analytics-class}
    \end{subfigure}

    \caption{Additional instructor-facing analytics in \drafttrace{}:
    (a) overall analytics for an individual submission,
    (b) longitudinal analysis of a student's writing across submissions,
    and (c) comparison with class-level writing patterns.}
    \label{fig:additional-instructor-analytics}
\end{figure}

Figure~\ref{fig:additional-instructor-analytics} presents additional instructor-facing analytics in \drafttrace{}, including an overall summary of an individual submission, a longitudinal view of the student's writing across submissions, and a class-level comparison for the assignment.

\end{document}